\documentclass[conference]{IEEEtran}
\IEEEoverridecommandlockouts
\usepackage{cite}
\usepackage{amsmath,amssymb,amsfonts}
\usepackage{algorithmic}
\usepackage{graphicx}
\usepackage{textcomp}
\usepackage{xcolor}
\usepackage{tabularx}
\usepackage{float}      
\usepackage{placeins}   
\usepackage{booktabs}
\def\BibTeX{{\rm B\kern-.05em{\sc i\kern-.025em b}\kern-.08em
    T\kern-.1667em\lower.7ex\hbox{E}\kern-.125emX}}
\begin{document}

 \title{Spatial Adaptation Layer: Interpretable Domain Adaptation For Biosignal Sensor Array Applications 
 \\ 
\thanks{This project was supported by UK Research and Innovation [UKRI Centre for Doctoral Training in AI for Healthcare grant number EP/S023283/1], the Imperial-META Wearable Neural Interfaces Research Centre and the Onassis Foundation.}
}


\author{\IEEEauthorblockN{Joao Pereira}
\IEEEauthorblockA{\textit{Department of Computing} \\
\textit{Imperial College London}\\
London, United Kingdom \\
jp2717@ic.ac.uk}
\and
\IEEEauthorblockN{Michael Alummoottil}
\IEEEauthorblockA{\textit{Department Bioengineering} \\
\textit{Imperial College London}\\
London, United Kingdom \\
mra21@ic.ac.uk}
\and
\IEEEauthorblockN{Dimitrios Halatsis}
\IEEEauthorblockA{\textit{Deparment of Bioengineering} \\
\textit{Imperial College London}\\
London, United Kingdom \\
dc23@ic.ac.uk}
\and
\IEEEauthorblockN{Dario Farina}
\IEEEauthorblockA{\textit{Department Bioengineering} \\
\textit{Imperial College London}\\
London, United Kingdom \\
\small{d.farina@imperial.ac.uk}}
}

\maketitle

\begin{abstract}
Machine learning offers promising methods for processing signals recorded with wearable devices such as surface electromyography (sEMG) and electroencephalography (EEG). However, in these applications, despite high within-session performance, intersession performance is hindered by electrode shift, a known issue across modalities. Existing solutions often require large and expensive datasets and/or lack robustness and interpretability. Thus, we propose the Spatial Adaptation Layer (SAL), which can be applied to any biosignal array model and learns a parametrized affine transformation at the input between two recording sessions. We also introduce learnable baseline normalization (LBN) to reduce baseline fluctuations. Tested on two HD-sEMG gesture recognition datasets, SAL and LBN outperformed standard fine-tuning on regular arrays, achieving competitive performance even with a logistic regressor, with orders of magnitude less, physically interpretable parameters. Our ablation study showed that forearm circumferential translations account for the majority of performance improvements.
\end{abstract}


\section{Introduction}
\IEEEPARstart{B}{iosignal} acquisition has powered a plethora of healthcare applications and wearable devices. Due to their complex structure, machine learning emerged as a promising method of leveraging biosignal information, such as movement intent classification from surface electromyography (sEMG) and electroencephalography (EEG) for human machine interactions (HCI) \cite{meta}.

Furthermore, sensor arrays enable the acquisition of spatially distributed information. Not only can this information be used for the inverse modeling of anatomically interpretable quantities, such as electrocardiography imaging (ECGi) \cite{ruben} or EMG decomposition \cite{emg-deep-learning}, but its regular structure also enables image-based processing \cite{csl} and deep learning approaches\cite{capgmyo} to be applied for these downstream tasks.

While aforementioned methods often yield high performance within the same recording session, performance is generally impractically low across sessions \cite{tackling-electrode-shift}. The displacement of sensors across sessions, referred to as electrode shift, is a known contributor in sEMG \cite{tackling-electrode-shift}, EEG \cite{eeg-shift}, and functional near-infrared spectroscopy (fNRIS) \cite{fnirs-shift} that degrades performance. Some studies have aimed to learn representations invariant to electrode shift \cite{joao}. While Meta Reality Labs has obtained remarkably low error across sessions and even subjects \cite{meta}, this result is due to a large dataset that can only be obtained through a resource intensive data collection effort, lacks interpretability, and only works on their own hardware, restricting it from external applications.

Beyond invariance learning, another proposed solution is calibrating with data from the new session. Image and signal processing methods have been used such as detecting anatomical landmarks for virtually reversing electrode displacement \cite{csl}. Other studies have attempted supervised domain adaptation techniques, such as the use of fine-tuning, progressive neural networks \cite{pnns}, and even model-agnostic meta-learning \cite{metalearning}. One promising approach introduces a two-step domain adaptation procedure, where an RNN is used for training on one session and a linear layer is used to map inputs to the input space of the previous session with frozen RNNs weights \cite{linear-layer}. 
However, either signal processing approaches are highly dependent on heuristic choices and independent preprocessing steps, or learning based approaches are non-interpretable and require large amounts of adaptation data, limiting their applicability in settings that require very short adaptation sessions.

\begin{figure}[t]
\includegraphics[width=0.9\linewidth]{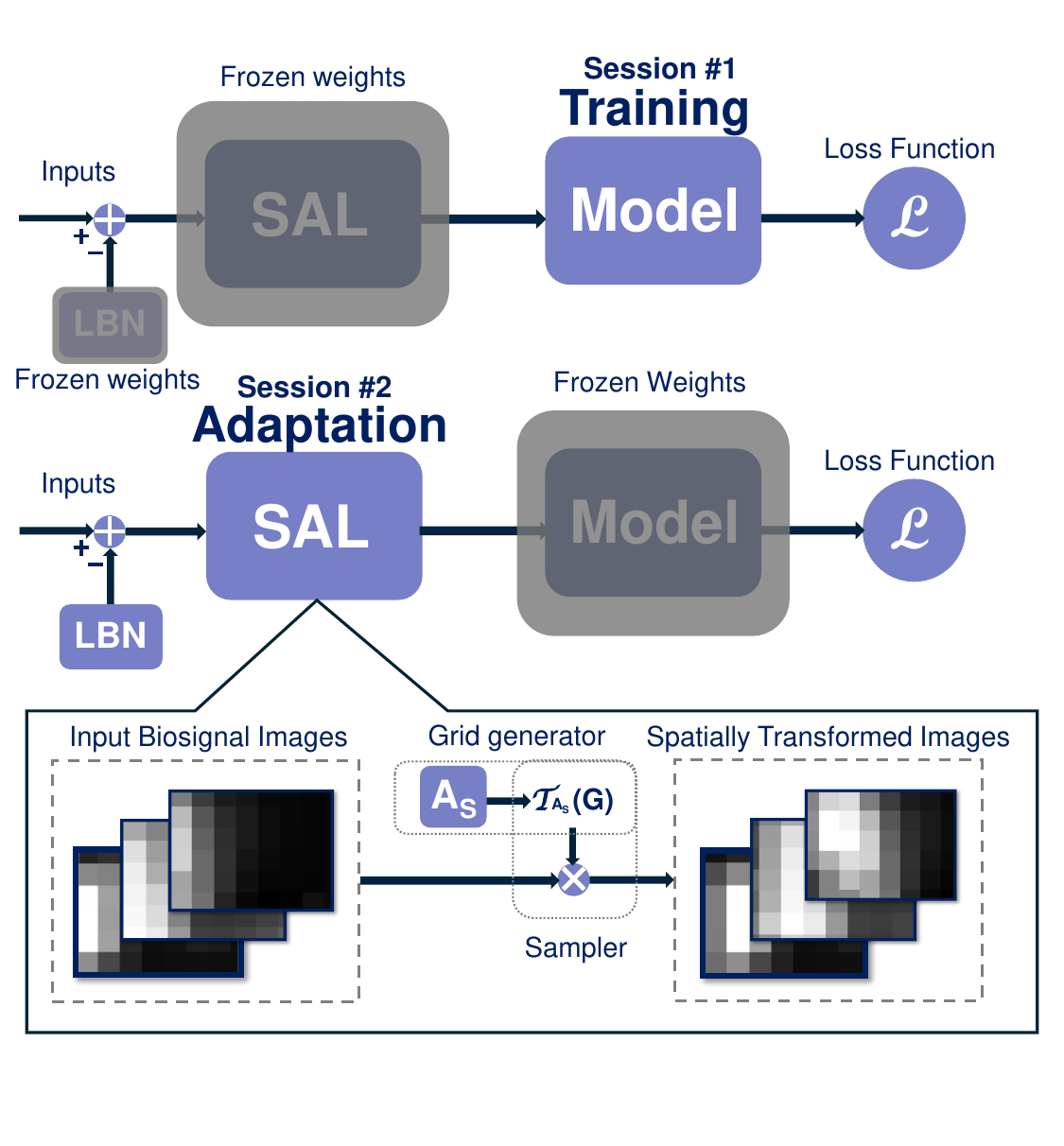}%
\caption{Schematic of SAL architecture and usage. SAL weights are frozen and the chosen model is optimized on session \#1. On session \#2, given a few examples, the model is frozen and the parameters of SAL and LBN are optimized for adaptation. SAL is similar to an STN module from \cite{spatial-transformers}, with the key difference that rather than modelling affine coefficients $A_{s}$ as a function of the input, $A_{s}$ are treated as learnable parameters.} 
\label{schematic}
\end{figure}

Therefore, in this paper, we introduce the Spatial Adaptation Layer (SAL). The philosophy behind SAL is that transforming biosignals from a new recording session back into the original spatial frame in which the original classifier was trained on would have a significant impact on the loss. Consequently, parameters of a single, unique affine transformation can be learned directly from the supervision loss during adaptation. On top of SAL, we also subtract biases from each biosignal channel that are learned during adaptation to account for baseline activity fluctuations. Inspired by the baseline normalization (BN) procedure from \cite{csl}, this method is referred as learnable BN (LBN). We evaluated our methods on two publicly available HD-sEMG datasets. For regular biosignal arrays, our interpretable method offered higher performance than standard fine-tuning. To summarize, as shown in Figure \ref{schematic}, we introduce:

\begin{enumerate}
    \item Spatial Adaptation Layer: which learns to directly mitigate spatial transformations applied to a regular biosignal array from the supervision loss
    \item Learnable Baseline Normalization: which learns to directly mitigate baseline activity fluctuations from the supervision loss 
\end{enumerate}

\section{Relevant Work}
SAL was inspired by and uses the differentiable re-sampling operator from Spatial Transformer Networks (STNs) \cite{spatial-transformers}.
While STNs compute coefficients of an affine transformation as a function of the given input (localization network), SAL treats these coefficients as learnable parameters, decomposed into separable transformations for interpretabillty, which are optimized through an adaptation session. Without a localization net, often a black box model, we bring further interpretability to the model. While similar in structure to the two-step adaptation introduced in \cite{linear-layer}, which prepends a linear layer instead of SAL, our method uses far fewer parameters, being more efficient and directly interpretable as spatial transformations to the signal array.

\section{Methods \& Datasets}

\begin{figure*}[ht!]
\centerline{\includegraphics[width=0.78\textwidth, clip, trim=20 75 0 10]{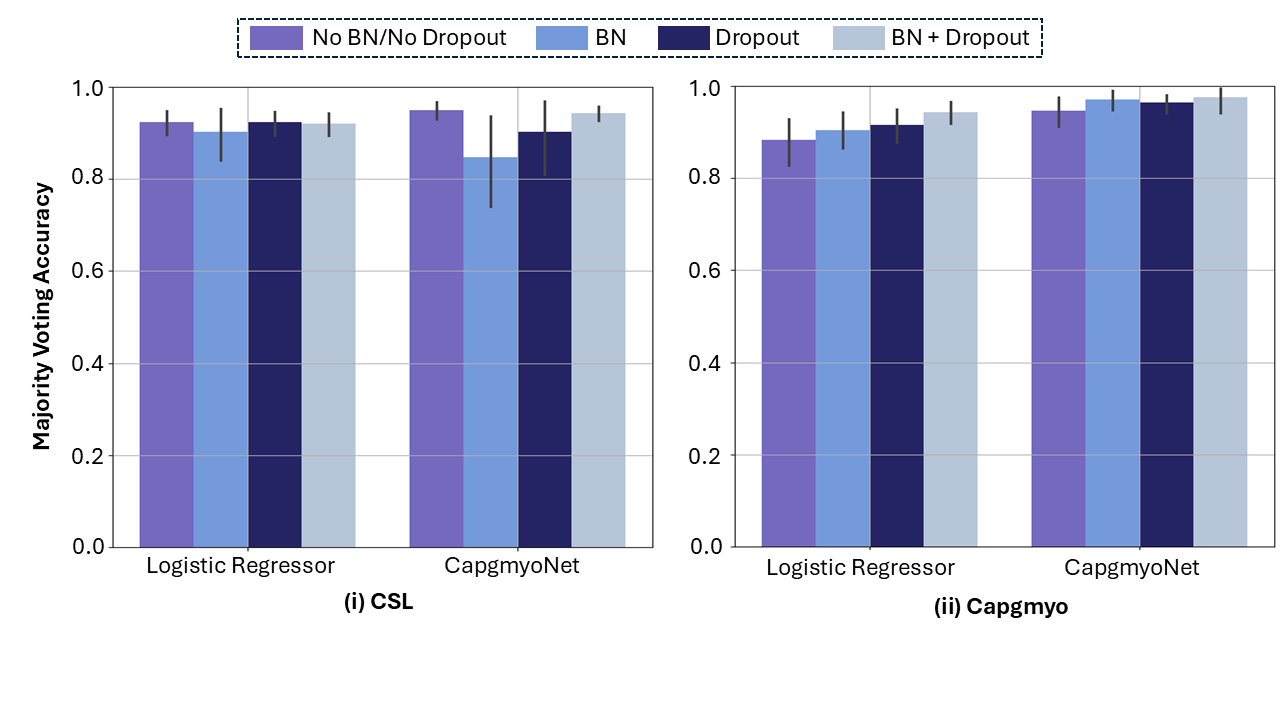}}
\caption{Majority voting classification accuracies for the intrasession protocol across different models and experimental conditions for \textbf{(i)} CSL and \textbf{(ii)} Capgmyo. For intrasession analysis, we train the classifier on all repetitions except one held out repetition per gesture, and obtain test accuracies on this held-out set. Across conditions, datasets and classifiers, high accuracies were obtained.}
\label{fig:intrasession}
\end{figure*}

\subsection{Spatial Adaptation Layer}
In the Spatial Adaptation Layer, seen in Fig \ref{schematic}, we treat each temporal slice of the given 2D regular array as a grayscale image. We used the sub-differentiable image sampling operator proposed in STNs \cite{spatial-transformers}. This is framed as resampling an image based on a set of sampling coordinates $T_{A_S}(G)$, given original grid coordinates of the image, $G$, under some affine transformation $A_S$. This can be seen in Eq. \ref{affine_transform} with $(x^s_i, y^s_i)$ elements of the sampling coordinates, and $(x^t_i, y^t_i)$ elements of the target coordinates in $G$ of the output image, $\forall i \in [1 \ldots HW] $ for an image of dimensions $H \times W$:
\vspace{-0.2em}
{\small
\begin{equation}\label{affine_transform}
    \begin{pmatrix}
        x^{s}_{i} \\
        y^{s}_{i}
    \end{pmatrix}
    = T_{A_S}(G_i) = A_S \begin{pmatrix}
        x^{t}_{i} \\
        y^{t}_{i} \\
        1
    \end{pmatrix}
    = \begin{pmatrix}
        \theta_{11} & \theta_{12} & \theta_{13} \\
        \theta_{21} & \theta_{22} & \theta_{23}
    \end{pmatrix} 
    \begin{pmatrix}
        x^{t}_{i} \\
        y^{t}_{i} \\
        1
    \end{pmatrix}
\end{equation}}

Given an input 2D biosignal temporal slice $U$, the resampling to an output $V$ for a single channel is defined in Eq. \ref{generic_sampling} for some sampling kernel $k$ with parameters $\Phi_x$ and $\Phi_y$:
\vspace{-0.3em}
{\small
\begin{equation}\label{generic_sampling}
 V_i = \sum_{n=1}^{H} \sum_{m=1}^{W} U_{nm} k(x_i^s - m; \Phi_x) k(y_i^s - n; \Phi_y)  
\end{equation}}

For billinear interpolation this can be written as:
\vspace{-0.3em}
{\small
\begin{equation}
    V_i = \sum_{n=1}^H \sum_{m=1}^W U_{nm} \max(0, 1 - |x_i^s - m|) \max(0, 1 - |y_i^s - n|)
\end{equation}}
The authors in \cite{spatial-transformers} show that the partial derivatives of the output image with respect to the input and sampling grid can be defined as:
\vspace{-0.2em}
{\small
\begin{equation}
    \frac{\partial V_i}{\partial U_{nm}} = \sum_{n=1}^H \sum_{m=1}^W \max(0, 1 - |x_i^s - m|) \max(0, 1 - |y_i^s - n|)
\end{equation}}
\vspace{-0.2em}
{\footnotesize
\begin{equation} \label{sub_grad}
    \frac{\partial V_i}{\partial x_i^s} = \sum_{n=1}^H \sum_{m=1}^W U_{nm} \max(0, 1 - |y_i^s - n|) \begin{cases} 
0,&  |m - x_i^s| \geq 1 \\
1, &  m \geq x_i^s \\
-1, & m < x_i^s
\end{cases}
\end{equation}}
yielding a sub-differentiable sampling mechanism, seen in Eq. \ref{sub_grad}. Gradients can flow through since $\frac{\partial x^s_i}{\partial \theta}$ and $\frac{\partial y^s_i}{\partial \theta}$ can be easily derived from Eq. \ref{affine_transform}, effectively allowing backpropagation to be applied. For more interpretable parameters to be learned, we further decompose $A_S$ into sub-transforms  whilst preserving differentiability in the operator. We do this by making use of tranformations in homogenous coordinates such that the general affine transformation $A$ is defined as Eq. \ref{affine_decompose} with $T$, $R$, $S_c$ and $S_h$ representing translation, rotation, scaling and shearing respectively.

\begin{alignat}{2}\label{affine_decompose}
A &=  S_h S_c R T \nonumber \\
  &=      \begin{pmatrix}
         1 & sh_{x} & 0 \\
         sh_{y} & 1 & 0 \\
         0 & 0 & 1
     \end{pmatrix}
     \begin{pmatrix}
         s_{x} & 0 & 0 \\
         0 & s_{y} & 0 \\
         0 & 0 & 1
     \end{pmatrix} \notag \\
   &\phantom{=}\begin{pmatrix}
         \cos{\phi} & -\sin{\phi} & 0 \\
         \sin{\phi} & \cos{\phi} & 0 \\
         0 & 0 & 1
     \end{pmatrix}  
\begin{pmatrix}
         1 & 0 & T_x \\
         0 & 1 & T_y \\
         0 & 0 & 1
     \end{pmatrix}
\end{alignat}

Hence, we can redefine $A_S$ as the submatrix of $A$ such that $A_S = A[I,J]$ where $I = \{1,2\}$ and $J = \{1,2,3 \}$. Given tensor \textit{slicing} and \textit{multiplication} are differentiable operations, gradients are preserved with no loss of information and can be defined with respect to the sampling grid coordinates by applying the chain rule (e.g. $\frac{\partial x^s_i}{\partial T_x} = \frac{\partial x^s_i}{\partial \theta} \frac{\partial \theta}{\partial T_x}$). This affine transformation and layer is parametrised by 7 learnable parameters. Ordinarily, one would expect such parameters to be learned through a similarity metric applied to images directly as in image registration. However exact correspondence may be ambiguous in different scenarios. Instead, parameters can be learned for any optimization based downstream task. Here, classification loss is backpropagated for multiple windows during adaptation, guiding the learning of these physically meaningful parameters to uncover a unique mapping from session to session.

\subsection{Datasets}
\textbf{Capgmyo \cite{capgmyo}:} Signals were acquired via eight differential, silver, wet electrode array evenly spaced around the circumference of the arm, resulting in 128 simultaneous channel recordings (irregular grid of 8x16) acquired at 1000 Hz. All experiments were carried out with data subset DBb, containing two recording sessions from subjects. Due to data corruption in the last subject’s recordings, this experiment was carried forward with the first 9 subjects. Each session consisted of 10 repetitions of 8 isotonic and isometric hand and finger gestures (G = 8) held for 3-10s. As in \cite{capgmyo}, only the central 1s interval of each gesture was considered in this study. 


\textbf{CSL Dataset \cite{csl}:} Bipolar recordings were acquired at 2048Hz via an electrode array with 192 electrodes with an inter-electrode distance of 10 mm and taking differences between consecutive samples, resulting in a regular grid of 7x24 (168 channels). This dataset contains recordings from 5 different subjects across 5 different days. For each recording, subjects performed 10 repetitions of 27 different gestures. To ensure that the labelled data corresponds to the desired movement, the activity segmentation from \cite{csl} was used to determine the active movement segment. The rest (idle) gesture was excluded.

\begin{figure*}[ht!]
\centerline{\includegraphics[width=0.79\textwidth, clip, trim=20 80 0 10]{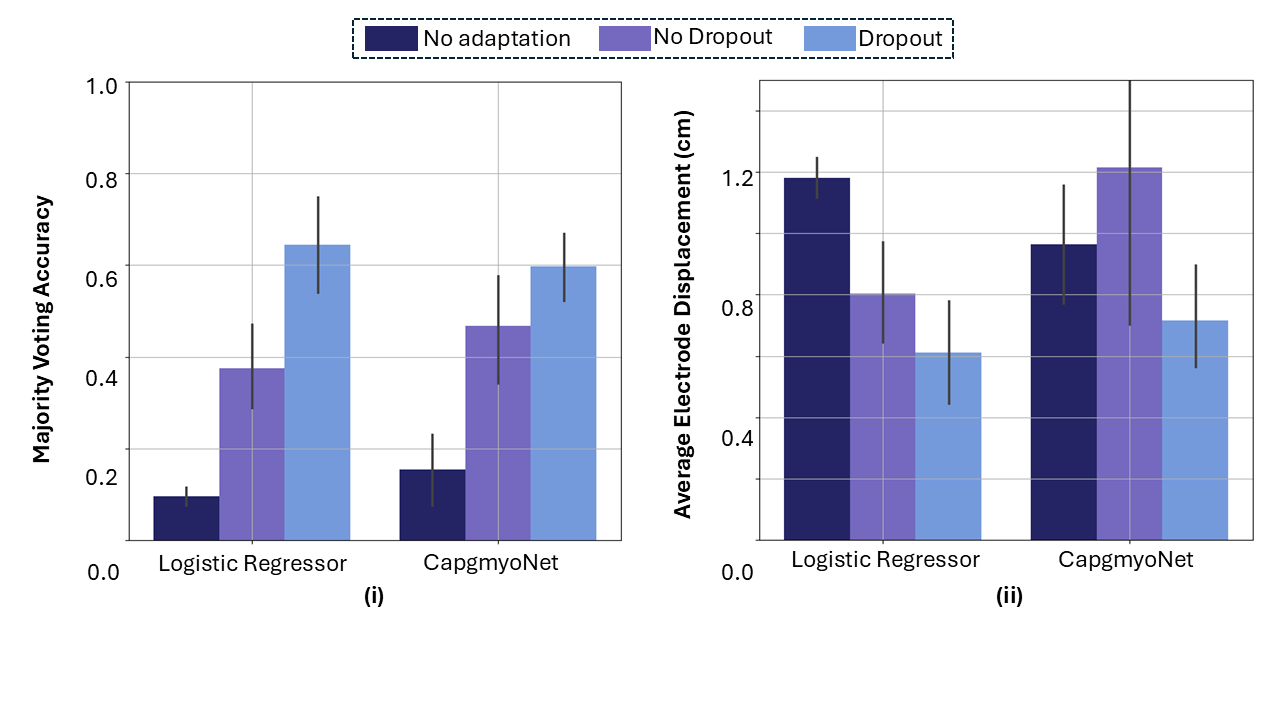}}
\caption{ Majority voting classification accuracies and average electrode displacement for the simulated spatial perturbations across different models and conditions for CSL data. With SAL and input dropout, distances are significantly decreased, improving performance from near chance-level to over 60\%.}
\label{fig:sims}
\end{figure*}

\subsection{Simulated Spatial Perturbations}
To evaluate the capability of SAL to mitigate the effect of spatial transformations between recording sessions of regular biosignal arrays, we applied spatial perturbations to held-out data of the same session the classifier was trained on. 
These spatial perturbations were parametrized in the same form as SAL, and acted as a controlled use-case where the ground truth spatial perturbation is known. For a given recording session of training data, one repetition of each gesture was randomly held out, and we applied a spatial perturbation to this held-out data by randomly sampling from $T_{x}, T_{y} \in [-2.0 cm, 2.0 cm] $, $\phi \in [-15.0^{o}, 15.0^{o}]$, $s_{x}, s_{y} \in [0.8, 1.2]$ and $sh_{x}, sh_{y} \in [-0.2, 0.2]$. 

As shown above, since we have access to the input grid coordinates ($G_{i}$) and output grid coordinates ($G_{o}$) given the implementation of SAL, we computed the average Euclidian distance between corresponding electrodes $d(G_{i}, G_{o})$. The more this metric decreases after applying SAL, the closer the transformed grid is to the original placement, acting as an interpretable metric of how well SAL addresses the true transformation applied. 

\subsection{Activity Images \& Learnable Baseline Normalization}
For a given time-window of $2T_{RMS} + 1$ samples, we can represent the biosignal tensor acquired as $\mathbf{U} \in \mathbb{R} ^{2T_{RMS} + 1 \times H \times W}$. Activity images were computed as in \cite{csl}, where the root mean square (RMS) of the time-window for each given channel is computed, resulting in activity image $\mathbf{U_{A}} \in \mathbb{R} ^{1 \times H \times W}$. To obtain an instantaneous estimate of RMS, the raw sEMG activity was squared, then convolved with a moving average (MA) filter of length $2T_{RMS} + 1$, introducing a delay of $T_{RMS}$ to the system, fixed to be 75ms.

Since, for a zero-mean signal, RMS acts as an estimate of channel standard deviation, baseline noise acts approximately as additive constants on $\mathbf{U_{A}}$. The baseline normalization (BN) procedure \cite{csl}, assuming a constant noise-variance, estimates baseline noise as $ \mathbf{N} = \mathbb{E}_{t}[\mathbf{U_{A}} | \mathbf{Y}=\text{rest}]$ from signals at rest, then subtracts from all gestures to obtain the processed input $\mathbf{U^{BN}_{A}} = \mathbf{U_{A}} - \mathbf{N}$. By applying BN independently to each recording session, it effectively removes the effect of baseline activity fluctuations across sessions. However, the estimation of $\mathbf{N}$ is highly dependent on the availability of consistently recorded rest activity, which is difficult to sustain in practice. Therefore, we propose learnable baseline normalization (LBN), by replacing the estimation of $\mathbf{N}$ with $\mathbf{B} \in \mathbb{R}^{1 \times H\times W}$, resulting in $\mathbf{U^{LBN}_{A}} = \mathbf{U_{A}} - \mathbf{B}$. By making $\mathbf{B}$ learnable, we have one additive constant per channel, enabling the model to optimally mitigate baseline fluctuations across session from the supervised loss. Note that while SAL is suitable for any regular biosignal array, LBN is bound to the processing of activity images or power spectral analysis as used in biosignals such as EEG and MEG.

\subsection{Preprocessing \& Classification}
As in \cite{capgmyo}, signals were processed with a digital bandpass filter (20-380 Hz, fourth-order Butterworth) and power-line interference was reduced using a digital band-stop filter (45-55 Hz, fourth-order Butterworth). To demonstrate the flexibility and simplicity of SAL and LBN, we integrated it into two different classifiers. The first is a multinomial logistic regressor (LogReg) trained with 2 epochs and batch size of 1024. Secondly, we implemented the SOTA sEMG image classifier from \cite{capgmyo}, which we refer to as CapgmyoNet (CpgmNet), trained with 1 epoch and a batch size of 128. As opposed to activity images, as recommended in \cite{capgmyo}, CapgmyoNet processed raw sEMG image frames on Capgmyo data. To prevent overfitting to noisy/corrupted channels during training, the effect of dropout at the input ($p=0.5$) was considered, probing the system to reduce dependencies on a small number of specific channels.

Given the findings from \cite{goyal}, we optimized the loss based on the sum across the batch, as opposed to the average, to leverage the linear scaling rule between learning rate and batch size and fix a learning rate for different batch sizes. For this to be valid, we also implemented a gradual warm-up learning rate schedule, where the learning rate was linearly increased from 1\% to 100\% of the base learning rate over the first training epoch. All classifiers were trained using the \textit{Adam} optimizer, cross-entropy loss, and a base learning rate of 0.05.

For fine-tuning, all parameters were trainable during adaptation. For spatial adaptation, as in Figure \ref{schematic}, only SAL and LBN parameters are frozen during classifier training. During adaptation, all model parameters are frozen except SAL and LBN parameters. Across models, the same hyperparameters in training were used during adaptation, with the number of epochs being multiplied by a factor 10 (number of repetitions per gesture), ensuring the same number of iterations as training.

For a fair performance comparison with previous studies, we consider the majority voting accuracy, defined as the mode of classification predictions in a given time-window. As done in \cite{csl, capgmyo}, we compute majority voting over for the full gesture repetition for CSL and 150ms windows for Capgmyo.

\section{Experimental Analysis}

\subsection{Intrasession Performance}
The first experiment involves obtaining the intrasession performance scores for the aforementioned classifiers on each of the datasets, acting as a practical upper-bound for intersession performance. For each intrasession experimental setting, 9 repetitions of each gesture were randomly selected from a recording session for training, and evaluated on data of the left-out repetition. As shown Figure \ref{fig:intrasession}, accuracies are high across models, with the optimal condition for each model being above 93\% accuracy. Interestingly, even a logistic regressor is able to achieve competitive performances, outperforming CapgmyoNet with Capgmyo data.

\begin{figure*}[ht!]
\centerline{\includegraphics[width=0.79\textwidth, clip, trim=20 75 0 10]{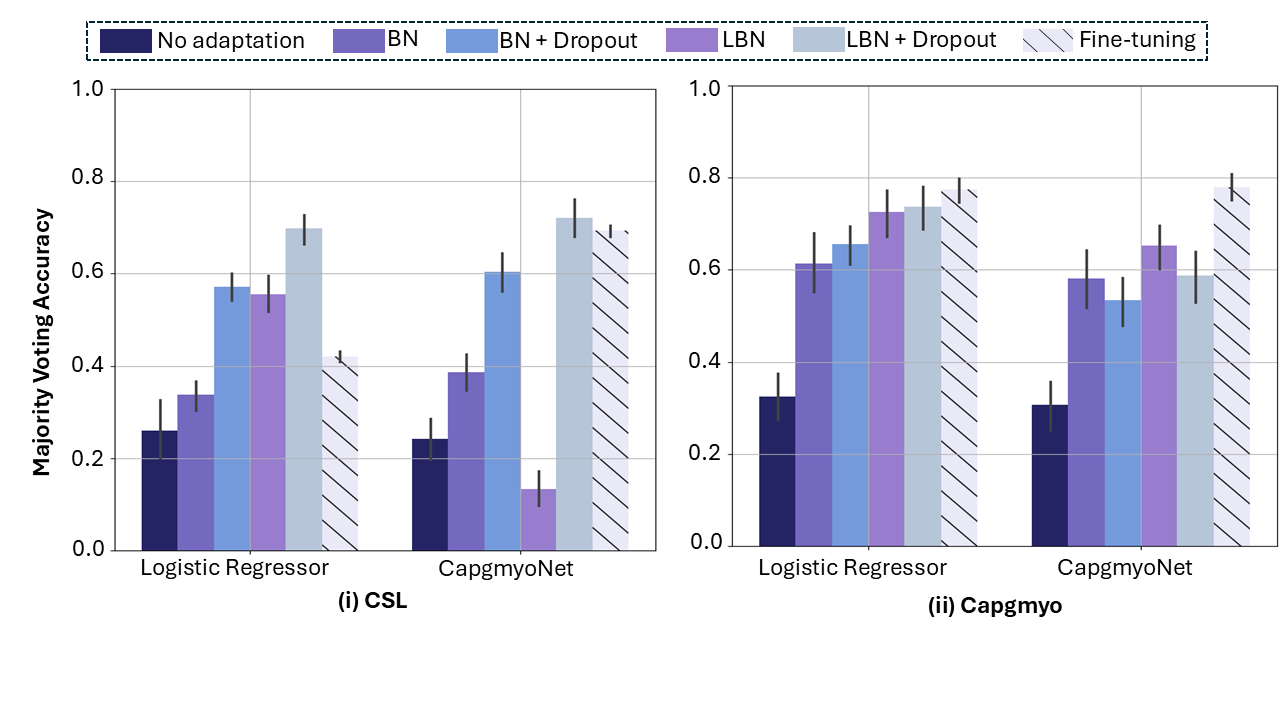}}
\caption{Majority voting classification accuracies for the intersession protocol across different models and experimental conditions for \textbf{(i)} CSL and \textbf{(ii)} Capgmyo. For intersession analysis, we train the classifier on one recording session, then adapt to the second session with one repetition per gesture, using the remaining data from the second session to obtain test performance. We then obtain test accuracies on this held-out set. While SAL + LBN seems to significantly outperform fine-tuning for CSL (regular grid), fine-tuning still performs better on Capgmyo data (irregular grid).}
\label{fig:intersession}
\end{figure*}

\subsection{Simulated Spatial Perturbations}
To evaluate the ability of SAL to account for spatial transformations, we investigate how well it can correct simulated spatial perturbations. This experiment used the same experimental protocol and data as in intrasession, but applied a randomly sampled spatial perturbation to the held-out test set. The models were then adapted and on this held-out set, on which we also obtain the final test accuracy and average electrode displacement ($d(G_{i}, G_{o})$). We evaluated SAL without LBN or BN, and only consider CSL data, as it uses a regular grid. As shown in Figure \ref{fig:sims}, across models, we see that the use of input dropout is pivotal to ensuring the classifier does not overfit to a small subset of channels, resulting in higher performances and significantly lower distance values. When using input dropout, both models were able to go from near chance level performance to over 60\% majority voting accuracy. This was further reflected by the significant decreases in distance values. These results support the use of SAL to correct for physical spatial transformations to biosignal arrays.

\subsection{Intersession Performance}
Intersession performance is measured as the accuracy from training on one session and testing on another session of the same subject. For each intersession experimental setting, for every permutation of 2 sessions for a given subject, the model was trained on the first session, adapted on one randomly sampled repetition of each gesture from the second session, and evaluated on the remaining 9 repetitions of each gesture of the second session. In our approach, SAL and LBN are optimized during an adaptation step, which is compared against fine-tuning (naive approach). We expected that since the data from Capgmyo is not a regular grid (variable interelectrode distance), assumptions made by SAL would distort the images and not be as beneficial towards performance, whereas the regular grids used in CSL make it the ideal SAL test case.

From Figure \ref{fig:intersession}, as expected, while all models achieved higher performance with fine-tuning for Capgmyo, for the regular grid data in CSL the optimal performance was achieved with SAL, despite using orders of magnitude less parameters. Furthermore, when using SAL, it can be clearly seen across models that the use of LBN significantly improved performance over traditional BN, highlighting the flexibility of learning the optimal baseline activity to subtract over estimating it directly from the data. As with simulated perturbations, input dropout proved vital when using SAL, consistently resulting in accuracy improvements of over 20\%, up to an astonishing improvement of almost 60\% for CapgmyoNet. The input dropout likely reduces its likelihood of overfitting to specific channels. Additionally, while SAL leaves the original learned representation intact, further experiments showed that after fine-tuning, performance on the original training set was less than 1\% higher than zero-shot intersession, indicative of catastrophic forgetting \cite{pnns}. Finally, while the CNN outperformed logistic regression, it is promising to see that with SAL applying domain adaptation at the input level, even traditional, more interpretable classifiers can achieve comparable performance to SOTA EMG-based classifiers.

\subsection{Model Ablation \& Interpretability Assessment}
Given the results from the intersession experiment, we explored the importance of different affine coefficients on the final performance for each model and dataset, using input dropout and LBN. Given that the signal array geometry and often orientation are consistent across experiments in both datasets considered, especially in sEMG, we expected that the translation parameters would be the most significant out of the 7 learnable coefficients. In addition, since electrode displacements along the circumference of the arm are most detrimental to sEMG classification performance as muscles are mostly distributed along the length of the forearm \cite{tackling-electrode-shift, joao}, we expected that corrections of such translations with SAL are likely to yield the largest performance improvements.

Thus, we run an ablation experiment of SAL with four different conditions: LBN (no affine), frozen translation coefficients, all SAL coefficients frozen except for translation, and keeping only the circumferential shift parameter learnable. As in the intersession experiment, performances reported use majority voting over the full active segment for CSL and 150ms window for Capgmyo, as in \cite{capgmyo}. Non majority-voted performances are reported in parentheses.

As shown in Table \ref{table:ablation}, while LBN alone accounts for significant improvements, SAL accounts for the majority of the improvements with only 7 learnable parameters. Furthermore, as expected, only keeping 2 learnable translation parameters results in significantly higher performance than the remaining 5 affine parameters. In line with past studies, the performance for only circumferential translation (1 affine coefficient) is nearly identical to the performance using both translation coefficients. Thus, SAL is able to drastically improve performance using few parameters closely associated with the physical cause of performance detriments. While these trends were observed for both datasets, non-translation affine parameters seemed to account for a larger proportion of performance with Capgmyo than CSL, as they are more useful for distortions introduced by the irregular geometry of Capgmyo grids.

\begin{table}[t!]
\caption{Ablation Experiments (Test Accuracy)}
\footnotesize
\begin{tabularx}{0.48\textwidth}{>{\centering\arraybackslash}X|>{\centering\arraybackslash}X>{\centering\arraybackslash}X>{\centering\arraybackslash}X>{\centering\arraybackslash}X}
\toprule
& LBN \hspace{20pt} (no SAL) & No translation & Translation only & Circumferential translation only           \\ \midrule
\textbf{CSL} \\
LogReg  & 0.433 & 0.497 & 0.647 & 0.644 \\
CpgmNet  & 0.526 & 0.592 & 0.723& 0.714 \\ \midrule

\textbf{Capgmyo} \\
LogReg  & 0.553 & 0.628 & 0.709 & 0.704 \\
CapgmNet & 0.492 & 0.530 & 0.577 & 0.607 \\ \midrule

\end{tabularx}
\label{table:ablation}
\vspace{-10pt} 
\end{table}

\section{Conclusion}
We have shown that SAL can be prepended to any differentiable model, from a logistic regressor to CNNs, and achieved higher intersession performances than even standard fine-tuning with orders of magnitude less parameters for biosignal arrays. In addition, LBN better accounted for baseline activity fluctuations over traditional BN making it particularly well suited for biosignal sensor arrays such as sEMG. Ablation experiments suggest that learned SAL coefficients are directly associated with the physical causes of the performance detriment. Interestingly, major improvements in sEMG systems by only accounting for circumferential translations suggest the possibility of applying the proposed approach to more practical 1D sensor arrays, such as the one in \cite{meta}.

Using one repetition per class, as in this study, is commonplace in sEMG literature \cite{metalearning}, with some studies using half \cite{linear-layer} or even all \cite{capgmyo} of the target data for adaptation. However, this becomes impractical for large number of classes. Thus, given that SAL and LBN apply domain adaptation at an input level, and electrode displacement and baseline fluctuations being invariant to class labels, future studies will explore whether our approach can adapt to one repetition of one (or a few) class(es), unlike present domain adaptation approaches.

\vspace{-5pt} 
\bibliographystyle{IEEEtran}
\bibliography{references}

\begin{thebibliography}{10}
\providecommand{\url}[1]{#1}
\csname url@samestyle\endcsname
\providecommand{\newblock}{\relax}
\providecommand{\bibinfo}[2]{#2}
\providecommand{\BIBentrySTDinterwordspacing}{\spaceskip=0pt\relax}
\providecommand{\BIBentryALTinterwordstretchfactor}{4}
\providecommand{\BIBentryALTinterwordspacing}{\spaceskip=\fontdimen2\font plus
\BIBentryALTinterwordstretchfactor\fontdimen3\font minus \fontdimen4\font\relax}
\providecommand{\BIBforeignlanguage}[2]{{%
\expandafter\ifx\csname l@#1\endcsname\relax
\typeout{** WARNING: IEEEtran.bst: No hyphenation pattern has been}%
\typeout{** loaded for the language `#1'. Using the pattern for}%
\typeout{** the default language instead.}%
\else
\language=\csname l@#1\endcsname
\fi
#2}}
\providecommand{\BIBdecl}{\relax}
\BIBdecl

\bibitem{meta}
\BIBentryALTinterwordspacing
D.~Sussillo, P.~Kaifosh, and T.~Reardon, ``A generic noninvasive neuromotor interface for human-computer interaction,'' 2024. [Online]. Available: \url{https://www.biorxiv.org/content/early/2024/07/23/2024.02.23.581779}
\BIBentrySTDinterwordspacing

\bibitem{ruben}
\BIBentryALTinterwordspacing
R.~R.-M. Serrano, S.~Velasco-Bosom, A.~Dominguez-Alfaro, M.~L. Picchio, D.~Mantione, D.~Mecerreyes, and G.~G. Malliaras, ``High density body surface potential mapping with conducting polymer-eutectogel electrode arrays for ecg imaging,'' \emph{Advanced Science}, vol.~11, no.~27, p. 2301176, 2024. [Online]. Available: \url{https://onlinelibrary.wiley.com/doi/abs/10.1002/advs.202301176}
\BIBentrySTDinterwordspacing

\bibitem{emg-deep-learning}
D.~Xiong, D.~Zhang, X.~Zhao, and Y.~Zhao, ``Deep learning for emg-based human-machine interaction: A review,'' \emph{IEEE/CAA Journal of Automatica Sinica}, vol.~8, no.~3, pp. 512--533, 2021.

\bibitem{csl}
\BIBentryALTinterwordspacing
C.~Amma, T.~Krings, J.~B\"{o}er, and T.~Schultz, ``Advancing muscle-computer interfaces with high-density electromyography,'' in \emph{Proceedings of the 33rd Annual ACM Conference on Human Factors in Computing Systems}, ser. CHI '15.\hskip 1em plus 0.5em minus 0.4em\relax New York, NY, USA: ACM, 2015, pp. 929--938. [Online]. Available: \url{10.1145/2702123.2702501">http://doi.acm.org/10.1145/2702123.2702501}
\BIBentrySTDinterwordspacing

\bibitem{capgmyo}
\BIBentryALTinterwordspacing
Y.~Du, W.~Jin, W.~Wei, Y.~Hu, and W.~Geng, ``Surface emg-based inter-session gesture recognition enhanced by deep domain adaptation,'' \emph{Sensors}, vol.~17, no.~3, 2017. [Online]. Available: \url{https://www.mdpi.com/1424-8220/17/3/458}
\BIBentrySTDinterwordspacing

\bibitem{tackling-electrode-shift}
\BIBentryALTinterwordspacing
L.~Hargrove, K.~Englehart, and B.~Hudgins, ``A training strategy to reduce classification degradation due to electrode displacements in pattern recognition based myoelectric control,'' \emph{Biomedical Signal Processing and Control}, vol.~3, no.~2, pp. 175--180, 2008, surface Electromyography. [Online]. Available: \url{https://www.sciencedirect.com/science/article/pii/S1746809407001012}
\BIBentrySTDinterwordspacing

\bibitem{eeg-shift}
H.~Kim and C.-H. Im, ``\BIBforeignlanguage{en}{Influence of the number of channels and classification algorithm on the performance robustness to electrode shift in steady-state visual evoked potential-based brain-computer interfaces},'' \emph{\BIBforeignlanguage{en}{Front. Neuroinform.}}, vol.~15, p. 750839, Oct. 2021.

\bibitem{fnirs-shift}
F.~Wang, M.~Mao, L.~Duan, Y.~Huang, Z.~Li, and C.~Zhu, ``Intersession instability in fnirs-based emotion recognition,'' \emph{IEEE Transactions on Neural Systems and Rehabilitation Engineering}, vol.~26, no.~7, pp. 1324--1333, 2018.

\bibitem{joao}
J.~Pereira, D.~Halatsis, B.~Hodossy, and D.~Farina, ``Tackling electrode shift in gesture recognition with hd-emg electrode subsets,'' in \emph{ICASSP 2024 - 2024 IEEE International Conference on Acoustics, Speech and Signal Processing (ICASSP)}, 2024, pp. 1786--1790.

\bibitem{pnns}
U.~Cote-Allard, C.~L. Fall, A.~Drouin, A.~Campeau-Lecours, C.~Gosselin, K.~Glette, F.~Laviolette, and B.~Gosselin, ``Deep learning for electromyographic hand gesture signal classification using transfer learning,'' \emph{IEEE Transactions on Neural Systems and Rehabilitation Engineering}, vol.~27, no.~4, p. 760–771, 2019.

\bibitem{metalearning}
\BIBentryALTinterwordspacing
X.~Fan, L.~Zou, Z.~Liu, Y.~He, L.~Zou, and R.~Chi, ``Csac-net: Fast adaptive semg recognition through attention convolution network and model-agnostic meta-learning,'' \emph{Sensors}, vol.~22, no.~10, 2022. [Online]. Available: \url{https://www.mdpi.com/1424-8220/22/10/3661}
\BIBentrySTDinterwordspacing

\bibitem{linear-layer}
I.~Ketykó, F.~Kovács, and K.~Z. Varga, ``Domain adaptation for semg-based gesture recognition with recurrent neural networks,'' in \emph{2019 International Joint Conference on Neural Networks (IJCNN)}, 2019, pp. 1--7.

\bibitem{spatial-transformers}
\BIBentryALTinterwordspacing
M.~Jaderberg, K.~Simonyan, A.~Zisserman, and K.~Kavukcuoglu, ``Spatial transformer networks,'' 2016. [Online]. Available: \url{https://arxiv.org/abs/1506.02025}
\BIBentrySTDinterwordspacing

\bibitem{goyal}
\BIBentryALTinterwordspacing
P.~Goyal, P.~Doll{\'{a}}r, R.~B. Girshick, P.~Noordhuis, L.~Wesolowski, A.~Kyrola, A.~Tulloch, Y.~Jia, and K.~He, ``Accurate, large minibatch {SGD:} training imagenet in 1 hour,'' \emph{CoRR}, vol. abs/1706.02677, 2017. [Online]. Available: \url{http://arxiv.org/abs/1706.02677}
\BIBentrySTDinterwordspacing

\end{thebibliography}
\end{document}